\documentclass{optica-article}

\journal{opticajournal} 

\articletype{Research Article}

\usepackage{float}
\usepackage{lineno}

\begin{document}

\title{The Integrated Forward-Forward Algorithm: Integrating Forward-Forward and Shallow Backpropagation With Local Losses}
\author{Desmond Y.M. Tang\authormark{1}}
\authormark{1}Peking University


\begin{abstract*} 
 The backpropagation algorithm, despite its widespread use in neural network learning, may not accurately emulate the human cortex's learning process. Alternative strategies, such as the Forward-Forward Algorithm (FFA), offer a closer match to the human cortex's learning characteristics. However, the original FFA paper and related works on the Forward-Forward Algorithm only mentioned very limited types of neural network mechanisms and may limit its application and effectiveness. In response to these challenges, we propose an integrated method that combines the strengths of both FFA and shallow backpropagation, yielding a biologically plausible neural network training algorithm which can also be applied to various network structures. We applied this integrated approach to the classification of the Modified National Institute of Standards and Technology (MNIST) database, where it outperformed FFA and demonstrated superior resilience to noise compared to backpropagation. We show that training neural networks with the Integrated Forward-Forward Algorithm has the potential of generating neural networks with advantageous features like robustness.
\end{abstract*}

\section{Introduction}
For the past decade, Deep learning has made significant progress on countless tasks and problems, with backpropagation as a key contributing factor to train the deep models developed. However, in aspects of biological plausibility, although the model of deep neural networks for artificial intelligence is initially inspired by biological neurons' structures, the training process of these deep neural networks using backpropagation throughout the whole neural network does not seem biologically plausible. Researchers have argued that the long range, or global, backpropagation of gradient information has little evidence in neuron science and violates many principles of the cortex learning found in neural science, like locality and onlinety.\\\\
The original Forward-Forward Algorithm is one of the approaches in biological plausible learning as an alternative of backpropagation to train neural networks initially proposed by Hinton(2022)\cite{The Forward-Forward Algorithm: Some Preliminary
Investigations}. The original Forward-Forward algorithm, or FF, uses two different forward passes to replace the forward-backward pass in backpropagation and uses most of the activations in the network's hidden layers as local losses to update the parameters while doing forward-forward passes.\\\\
There have been some limitations and unanswered questions for the original Forward-Forward algorithm. On the evaluation of performance and application range, the original method of Forward-Forward Algorithm shows limited performance in training speed and accuracy. Besides, from the experiments we demonstrate, the original FF does not scale well to deeper neural networks compared with BP. In practice, literature only contains implementable code for a limited kind of network structure, DNN, according to our knowledge. In methodology, there have also been questions about the relationship between Backpropagation and Forward-Forward and whether we can combine the two to get a better training algorithm.\\\\
In response to these questions, we propose an integrated version of FF with shallow Backpropagation, or the Integrated Forward-Forward Algorithm, which is motivated by both FF and the use of local losses. IntFF harnesses the design of multiple local losses to decompose the training target and to avoid the process of passing gradients throughout the whole network structure. Then it uses shallow backpropagation of just 1-3 layers to adjust weights in some local hidden units, allowing more complex network structures with a range of mechanisms compared with the original FF. \\\\
The Integrated Forward-Forward Algorithm is a gradient-based algorithm for the training process of neural networks. Compared with the renowned algorithm of backpropagation calculating all the gradients of the trainable parameters using a global loss, this algorithm uses multiple local losses for different parts of the neural network instead. IntFF algorithm replaced the forward-backward mechanism in backpropagation with two different forward passes. Each training step, a positive data and a negative data is fed into the network, causing the network to have different activations in its hidden layer neurons and updating the parameters using these activations.\\\\
The parameters of the network will be updated while doing these forward passes according to some local losses. More specificly, the parameters are divided into a few parts, or hidden units, with each part attached to a local loss function, and the parameters in this hidden unit will be updated using the gradients calculated from optimizing this co-responding local loss using shallow backpropagation. \\\\
Unlike in normal situations where backpropagation can pass gradient information throughout the whole neural network with a large number, sometimes hundreds, of layers, these shallow backpropagations through only less than a handful of layers does not necessarily violate the highly locality characteristic for the cortex learning procedure. \\\\
After the training process, the network will behave differently when given different types of data, positive or negative. More specifically, it is encouraged by the training algorithm to have larger activation values when given positive data and smaller activation values when given negative data. Later on when making predictions, with proper technics to capture the differences, we can use the trained network to accomplish a few classification tasks even though the network does not necessarily contain an output layer.\\\\
Networks trained with IntFF should be able to accomplish various tasks with different settings. In this specific paper, we applied the method to do the MNIST classification problem, and we defined the local losses to be the mean squares of activations of a few selected hidden neurons groups. To predict a classification label for an image with the trained neural network, one can use different labels separately to calculate the activations, or local losses, of the trained network and use the label with the highest activation as the predicted label. One can also train another extral linear model attached to the neural network with an output layer indicating labels using this neural network trained with IntFF as useful representations.\\\\
Carrying out experiments with IntFF, we found that the networks trained with IntFF shows a higher resilience under noisy conditions and achieved better results compared with neural networks of the same size and structure trained with backpropagation, which shows that the IntFF algorithm has the potential of creating neural networks of better robustness and possibly other beneficial characteristics. We also provide useful insights on IntFF and the original FF beyond these experimental results. \\\\
\begin{figure}[ht]
\centering\includegraphics[width=13.8cm]{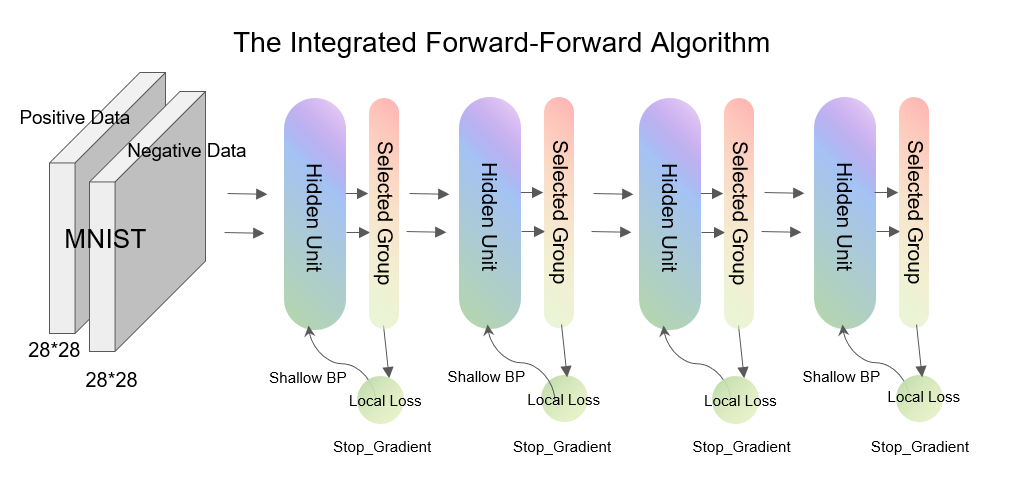}
\caption{IntFF: the Integrated Forward-Forward algorithm.}
\end{figure}
\section{Related Works}
Ever since the birth of neural networks, how to design algorithms to train these neural networks has aroused major interest in the feild of machine learning. There already exist review papers that systematicaly introduce the major approaches and progresses in biologically plausible deep learning\cite{A review of learning in biologically plausible spiking neural networks}\cite{Theoretical Neuroscience: Computational and Mathematical Modeling of Neural Systems (Computational Neuroscience Series)}. Here are some related subtopics that are most related to the Integrated Forward-Forward Algorithm. 
\subsection{The Idea of Two Forward Passes}
The idea of two forward passes passing a positive data and a negative data through the network to train the network is initailly mentioned by \cite{The Forward-Forward Algorithm: Some Preliminary
Investigations}replacing backpropagation by two forward passes. The original Forward-Forward Algorithm is inspired by the Boltzmanm machines\cite{Learning and relearning in
boltzmann machines} and by contrastive learning algorithms\cite{Noise-contrastive estimation: A new estimation principle for
unnormalized statistical models}. It constitutes a new candidate for biologically plausible learning. The ideas has also been applied later by \cite{Graph Neural Networks Go Forward-Forward}to Graph Neural Networks on the task of graph property prediction. Although the use of forward-forward algorithm has the potential of challenging the backpropagation and having a huge impact on the deep learning community, further works harnessing the forward-forward algorithm seems limited in the literature to our knowledge so far.
\subsection{Biologically Plausible Learning}
The Integrated Forward-Forward Algorithm is related with the goal of biologically plausible learning. Many biological plausible learning approaches uses perturbation in various types,\cite{The organization of behavior: a neuropsychological theory},\cite{Synaptic plasticity: taming the beast}. There have also been approaches about spiking neural networks with backpropagation in the literature.\cite{Learning in spiking neural networks by reinforcement of stochastic synaptic
transmission.}\\\\
There have been many perspectives arguing that backpropagation is biologically implausible with its properties of global loss function, non-local plasticity and update locking\cite{Hebbian Deep Learning Without Feedback}. On the contray, biological plausible learning algorithms should fit the characteristics of locality, online plausibility, and short range passes of information
\section{The Integrated Forward-Forward Algorithm}
\subsection{The Forward-Forward Mechanism}
The Forward-Forward mechanism is initially introduced by Geoffrey Hinton in his paper, "The Forward Forward Algorithm: Some Preliminary Investigations"(2022)\cite{The Forward-Forward Algorithm: Some Preliminary
Investigations}. From a calculative perspective, it replaces the forward-backward mechanism in the backpropagation procedure with two different forward passes. \\\\
In each pass, each layer of the neural network will calculate the linear transformation and activation functions sequentially. Then a layer-wise mean square will be calculated to represent the activation level of the neural network layer which can also be considered as an indicator of how "good" the passed data is. The mean square will then be used as a layerwise optimization target for the training process. After training, this layer-wise mean square will be used to compute a goodness function that can then be used for the prediction of classification labels when doing predictions.\\\\
It is worth mentioning that, if one wants to formulate the training procedure of FF as a optimization problem, the gradients calculated using the original Forward-Forward algorithm is the same as the gradients calculated using multiple local losses.\\\\
Further more, optimizing these local losses is the same as optimizing a global goodness function with the stop\_grad operation activated for each local loss. This goodness function is not directly the optimization target to calculate the gradients needed for updating our model parameters if one wants to derive accurate gradients from it. But it can be considered as the target using an engineering technic of stop\_gradient provided by some deep learning libraries like Pytorch or TensorFlow. Noticing that each $y_j$ are computed by $y_j=\Sigma_i w_{ij}*y_i+b_j$in the original FF where the indice $i$ takes value of the indices of the former layer, when calculating the gradients of $p(positive)=\sigma(\Sigma_j y_j^2 -\theta)$ by the parameters $w_{ij}$ and $b_j$ with stop\_gradient mechanism, it is the same as regarding all the $y_j^2$(s) as local losses and compute the accurate gradient of these layer-wise local losses by the weights and biases in the same layer. 
\begin{figure}[ht]
\centering\includegraphics[width=7cm]{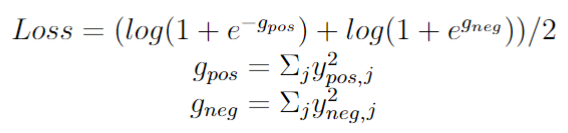}
\caption*{This is the loss function FF and IntFF use to update the parameters. $y_j$ stands for the activations in the neurons. $g_{pos}$ and $g_{neg}$ stand for the goodness for these two data points. When viewed as local losses, the sum takes all $j$(s) in a selected group of neurons and use shallow backpropagation to calculate the gradients in the hidden units. When viewed as a global loss, the sum takes all $j$(s) in the whole network and use backpropagation with stop\_grad to calculate all the parameters.}
\label{5}
\end{figure}
\begin{figure}[ht]
\centering\includegraphics[width=5.5cm]{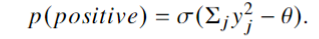}
\caption*{The likelihood function used to predict labels when the neural network is trained for both FF and IntFF. Here, $p(positive)$ stands for a kind of likelihood for the passed data to be a positive data representing a measure of goodness, $\sigma$ stands for the logistic function, $y_j$ stands for the activations of each neurons in the network, $j$ is the indice of the neuron, enumerating the whole neural network, and $\theta$ stands for a fixed threshold which is set to be $1.5$ for the whole process.}
\label{4}
\end{figure}
\subsection{The Integrated Forward-Forward Algorithm}
 Because gradients from FF can be viewed as gradients from local losses, it is natural to generalize the algorithm using local losses and more complex layers before the local losses. We propose IntFF as a more generalized version of the Forward-Forward algorithm, also using two Forward-Forward passes of positive data and negative data but with more complex local loss designs and co-responding layer structures defined as hidden units.\\\\
The architecture of our method is as given in the diagram, it sets up multiple local losses inside the neural network and uses shallow backpropagation to calculate the derivatives in the hidden units attached to the local losses. We call them hidden units 
because in computation, the local losses are only calculated from the selected neurons, or selected groups of neurons, not including the hidden units. And in the prediction of labels, we also use the activations of the selected neurons as an indicator for the goodness of the data, not including the activations in the hidden units. The only computational operations these hidden units are involved in include the forward pass procedure when given a data point to the neuron network and the gradient decent procedure to adjust the weights as a necessary part of our learning algorithm. These hidden units can be single neurons or groups of connected neurons that share the same local loss structure.\\\\
Considering the numerical fact that encouraging the activations to be large when given positive data can result in some numerical overflow when training, each hidden unit will first normalize the inputs from the former layer before operating the operations in the hidden unit. Thus the optimization target can be stated as maximizing the activations of the selected groups of neurons, as local losses, subjected to normalized inputs, separately in all the hidden units.
\subsection{Training Networks with Integrated Forward-Forward}
In the training process, each training step a positive data and a negative data will be passed to the network structure. Then the local loss functions will be calculated using the activation functions of a few selected neuron groups. Then these local losses will be used to update a group of parameters attached with these local losses separately. The training process encourages the neuron network to have higher activations when given positive data and to have a lower activation when given with a negative data. The training process does not involve a comparing procedure between predicted labels and true labels, instead, the true labels are overlayed on the training images as positive data points as a whole. \\\\
When a positive data is feed to the network for training, the parameters will be adjusted to minimize their co-responding local losses. When a negative is feed to the network for training, the parameters will be adjusted to maximize their co-responding local losses. Here, the local losses can be considered as a kind of negative activation level and should be optimized in the contrary direction. The process can be better demonstrated with our pseudocode.\\\\
One question remaining unanswered to the training process is that how can we get a good source of negative data. It is worth noticing that this negative data can not be randomly given. It should be properly selected instead.\\\\
For example, complete Gaussion noises are not capable to be the negative data source. If one uses complete Gaussian noises as the source for negative data, we won't be getting good enough neural networks from training. This is because Gaussian noises doesn't posses enough information to be considered as a proper contrast to the positive data to train the neural network. A general understanding about what kind of negative data can be considered as a good negative data source is that compared with the positive data, the negative data should have similar low-level features but different high-level features.\\\\
In practice, when doing classification problems, one can use the data with the correct label as positive data and use the data with randomly shuffled label as negative data. This ensures that the negative data source has very similar, or generally the same, lower level features as the positive data source and very different semantic features as the higher level feature required for the learning procedure.\\\\
There are also other methods for the creation of negative data, Hinton(2022) mentioned a kind of hybrid method to create the source of negative data by multiplying two different data points with an artificial mask and adding the two results up to get the negative data source of similar lower level features. There are also methods in contrastive learning suggesting that we can use rotation or noisy perturbation to get the demanded negative data. What kind of negative data works best for the Integrated Forward-Forward Algorithm still remains unknown. We believe that the choice of randomizing labels as the source of negative data is convenient to implement and commonly usable when dealing with classification problems.
\begin{figure}[ht!]
\centering\includegraphics[width=13.8cm]{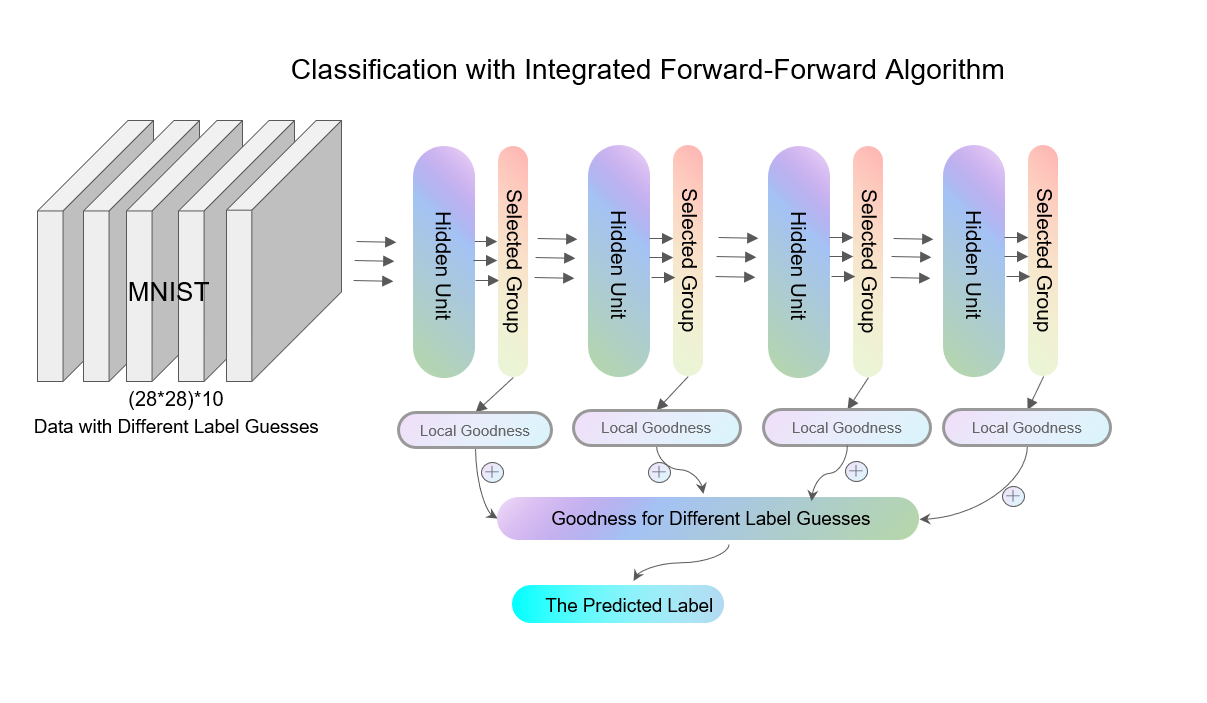}
\caption{Using the trained neural network to do classifications for MNIST}
\label{2}
\end{figure}
\subsection{Classification with the Integrated Forward-Forward}
When doing classification, unlike common neural network approaches, the Integrated Forward-Forward Algorithm does not necessarily include an output layer. Similar to the original Forward-Forward Algorithm, it does the classification task by calculating the sum of the activations of selected groups of neurons when given different labels to the input image and finding out which input assignment achieves the highest activation.\\\\
Compared with the original Forward-Forward Algorithm by Hinton, our method achieves better performance in training time and prediction accuracy and supports a larger range of neural network architectures with only slight changes in engineering. If you want to testify our method on a new network structure, it only takes about a handful of coding lines of change which is of significant engineering convenience compared with the Matlab version of the original FF.
\begin{figure}[ht!]
\centering\includegraphics[width=13.8cm]{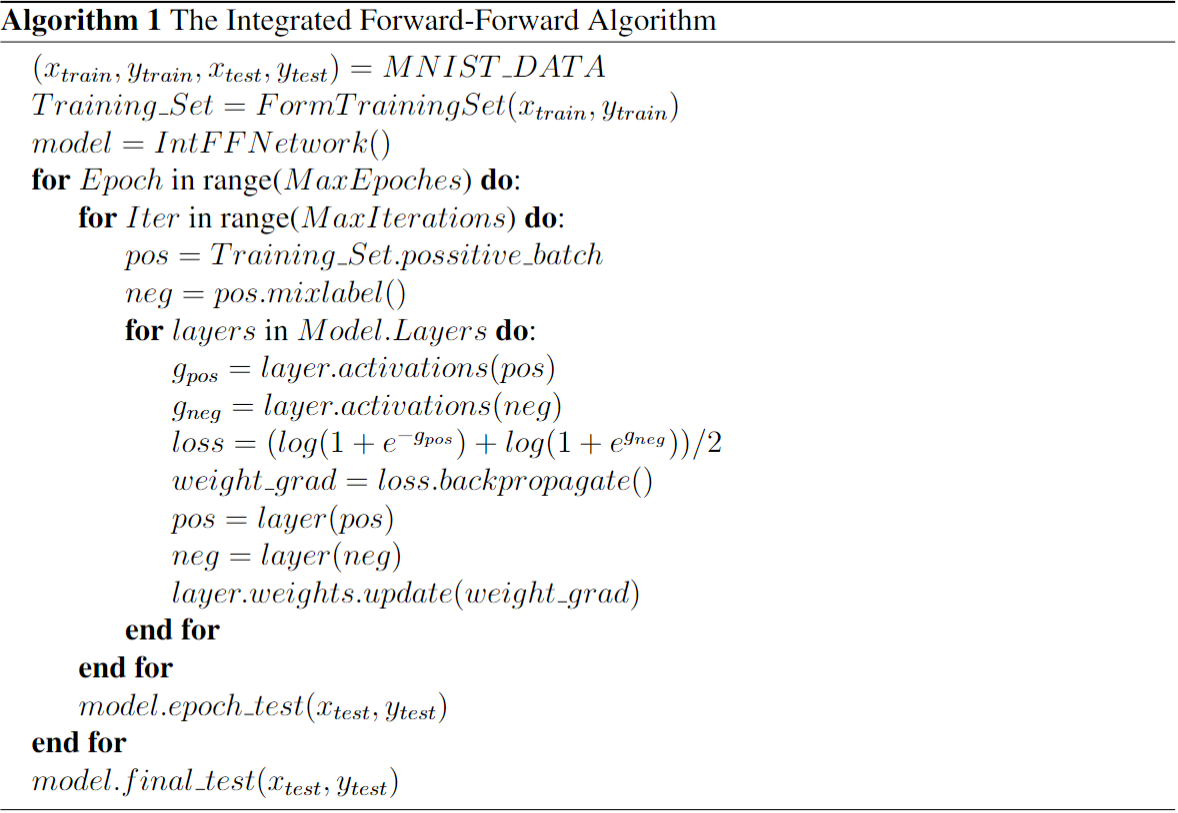}
\caption*{}
\label{1}
\end{figure}
\subsection{Why can IntFF Fit the Training Data}
Unlike many other deep learning approaches toward the classification problem, the Integrated Forward-Forward Algorithm and the original Forward-Forward Algorithms don't necessarily possess output layers. How such leaning algorithms can harness the training data to update its parameters is also different from deep learning methods with output layers to represent the predicted probability of labels.\\\\
In the training process, the dynamics of parameters are designed to encourage the neural network to have larger activations when given positive data and to have lower activations when given negative data. If the training process goes smooth, the neural network is going to behave differently given with different types of data. Even though this difference is not as explicit as situations where we have output layers, when we can enforce a neuron's activation to explicitly represent the likelihood of the label, if such difference can be learned from the dataset and detected using one way or another, the network can still benefit from the dateset. For example, after the training procedure, if one wants to apply the trained network to a classification problem, one can calculate the activations and compare as a way of detecting the concept by detecting some characteristics in the trained networks. One can also train another shallow softmax layer to output the predicted label as another way of detecting the concept of label by detecting some characteristics of the representations provided by the trained network.

\section{Experimental Results with MNIST}
\subsection{The MNIST Dataset}
The MNIST (Modified National Institute of Standards and Technology) dataset is a large database of handwritten digits that is widely used for training and testing in the field of machine learning. MNIST contains 70000 images of digits from 0-9 intotal, with each $28*28$ pixels. We implemented our method to the MNIST dataset and got 2.01\% of error rate with a really small network structure. We also made detailed comparison between the three training algorithms subjected to certain specific configurations.
\subsection{Using Images with Random Labels as Negative Data}
We've seen two different methods to create negative data source with MNIST in the context of Forward-Forward algorithm. One is to use real image with randomized label as negative data. Another one is to use a mask method hybriding different images introduced by\cite{The Forward-Forward Algorithm: Some Preliminary
Investigations}. We chose to use the negative data with random labels because this choice can be easily generalized to other image classification datasets compared with using hybrid images and easier to get.
\subsection{Analysis of the Performance}
In performance, IntFF showed competitive results compared with the original FF and BP. Compared with FF, IntFF shows better results when the network structure is larger and in networks with convolution and attention mechanisms. Actually, the original Forward-Forward algorithm is uneasy to be applied with such mechanisms directly as convolution operatiosn often results in multiple images with multiple channels and thus experimentally causing large hidden layers in its context without proper pooling or other methods to avoid this situation. Compared with BP, IntFF showed competitive results in most cases without noise and showed even better results under noisy situations.\\\\
It is worth mentioning that as newly developed biologically plausible algorithms the Forward-Forward and the Integrated Forward-Forward both shows promising results in image classification. Even though they can not beat backpropagation currently, they are still worth further exploration. For further implementation of developing, one can use our open-source python coding environment at https://github.com/Desmond-YM-Tang/The-Integrated-FOrward-Forward.
\begin{figure}[htbp]
\centering\includegraphics{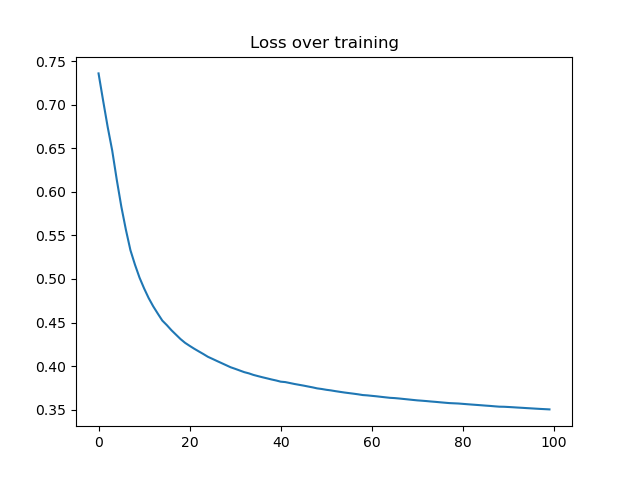}
\caption{The Change of the Mean Value of the Local Losses in IntFF Through Training. The x-axis indicates the num of training epochs.}
\end{figure}
\begin{table}[htbp]
\centering Training Results with the uncorrupted MNIST
\begin{tabular}{|c|c|c|c|c|}
\hline
Algorithm & Network Type & Network Size & Training Time & Testing Accuracy \\
\hline
IntFF & Dense & (784,(100,50),(30,10)) & 3081s & 94.04\% \\
\hline
FF & Dense &  (784,100,50,30,10) & 3523s & 94.64\% \\
\hline
BP & Dense &  (784,100,50,30,10) & 3086s & 96.07\% \\
\hline
IntFF & Dense & (784, (200, 200, 200), (50, 50)) & 7103s & 98.00\% \\
\hline
FF & Dense & (784, 200, 200, 200, 50, 50) & 7126s & 94.48\% \\
\hline
BP & Dense & (784, 200, 200, 200, 50, 50, 10) & 5300s & 97.27\% \\
\hline
BP & Convolution & (784,32C(3,3),64C(3,3),128,48,10) & 8291s & 97.71\% \\
\hline
\end{tabular}
\caption{A comparison among Integrated Forward-Forward, the original Forward-Forward and Backpropagation on the task of MNIST classification. IntFF stands for the Integrated Forward-Forward. FF stands for the original Forward-Forward. BP stands for Backpropagation. The training procedure used the same Adam optimizer provided by tensorflow and the same random seed for reproducability. Results with similar configurations are placed together to make better comparison. C(3,3) means the convolutoin kernels have the size of (3,3).}
\label{table:2}
\end{table}
\section{The Promising Robustness of the Integrated Forward-Forward}
\subsection{Experiments with Noisy Data}
Since IntFF is another approach to train neural networks compared with backpropagation, there might be some beneficial features for neural networks trained with IntFF share compared with BP. We tried this method with an artificially designed noisy dataset and found that IntFF often results in better robustness under noisy conditions compared with Backpropagation.\\\\
We intentionally corrupted the true MNIST dataset with three types of noises, as shown in the diagram, and then used IntFF, FF, BP to train neural networks with similar scale all with proper early stopping to prevent the network from overfitting into the noisy dataset. The results are shown in the tabula. It turns out that IntFF can give out rather good neural network training results, 
 about 5.5\% error rate, with dataset that is generally corrupted with a percentage of 75\% noisy data of different types. Neural Networks trained with backpropagation showed to be less competitive under such conditions.\\\\
This result as a by-product of this paper shows that the Integrated Forward-Forward algorithm and the original Forward-Forward algorithm has the pottential to generate networks with better robustness compared with the neural networks trained with backpropagation.
\subsection{Possible Explanations for the Robustness}
We think that this feature of robustness comes from a decentralized evaluation of the performance of neural network and a contrastive learning procedure. \\\\
For the decentralized evaluation, facing noisy conditions, although the real noise comes from the dataset, one can equivalently interpret the noise as coming from the losses and backpropagated to the weights. With multiple local losses, these noises might have contrary effects to the network parameters, thus leading to less randomized noises in gradients. Having only one global loss function which is affected by the noise might cause more misleading gradients to the training process and that might be one possible explanation for the robustness gap.\\\\
Also, with the contrastive learning paradigm, although the training images are corrupted with data, it can still form a contradict with the negative data created with it and make positive contribution to the learning algorithm. Using backpropagation without such contrastive mechanism might be the reason for less robustness.
\begin{figure}[htbp]
\centering\includegraphics[width=12cm]{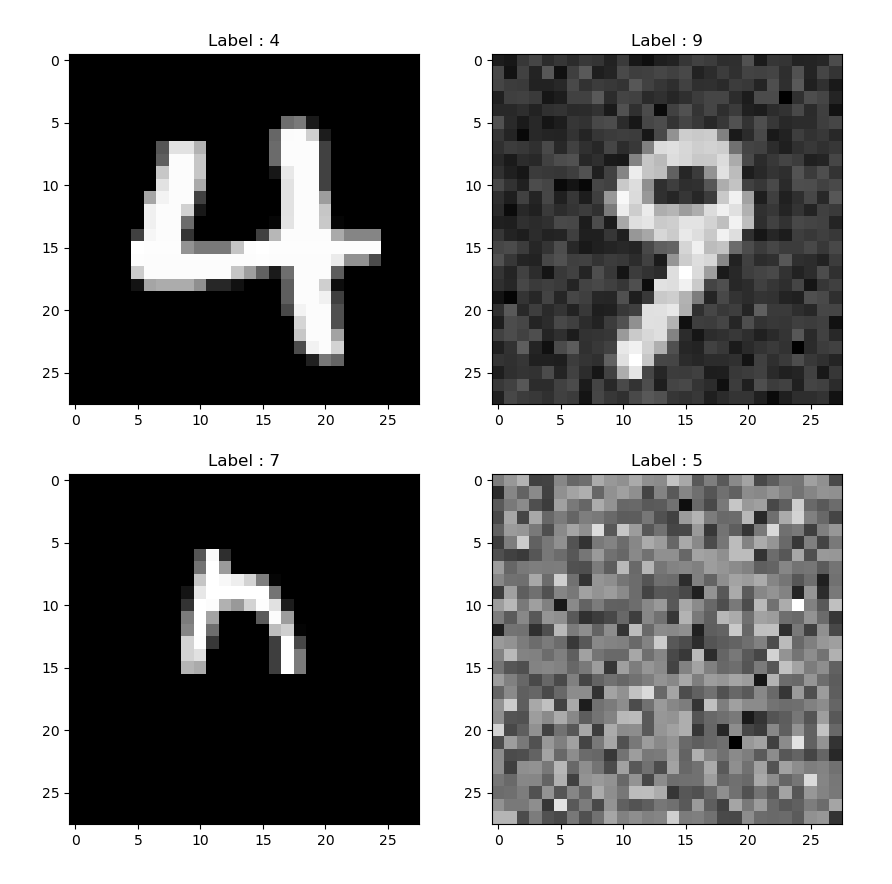}
\caption{The artificially corrupted dataset with a proportion of 25\% each type. The first type is the uncorrupted data. The second type is the authentic data corrupted with a Gaussian noise. The third type is the authentic data corrupted with a random number of unloaded pixels. The last type is completely Gaussian noise with random labels. It is a handcrafted noisy dataset to testify the robustness for these three methods. The testing procedures are still operated on the authentic teasting set.}
\end{figure}

\begin{table}
\centering Training Results with Noisy Training Data
\begin{tabular}{|c|c|c|c|}
\hline
Algorithm & Network Type & Network Size  & Testing Accuracy \\
\hline
IntFF & Dense &   (784, (100, 100), (100, 100))& 94.38\% \\
\hline
FF & Dense &    (784,100,100,100,100) & 90.78\% \\
\hline
BP & Dense &  (784,100,100,100,100,10)  & 93.83\% \\
\hline
IntFF & Dense &   (784,(100,100),(100,100),10) & 94.12\% \\
\hline
\end{tabular}
\caption{A comparison on the task of MNIST classification with the artificially corrupted dataset. The training procedure used the same Adam optimizer provided by tensorflow and the same random seed for reproducability.}
\label{table:1}
\end{table}
\section{Relationship Between the IntFF and the Original FF and BP}
In classification problems explored in this paper, the Integrated Forward-Forward algorithm can be considered as a generalization for both the original FF and the Backpropagation in the process of training neural networks. It is worth mentioning that all of these three training algorithms are gradient-based, and with certain specifications, the gradients calculated with the IntFF can be exactly the same with the gradients calculated using FF or BP with stop\_grad mechanism.
\subsection{Relationship Between the IntFF and the Original Forward-Forward Algorithm}
In comparison with the original Forward-Forward algorithm, IntFF broadens the network structures that can be trained using two contrastive forward passes of positive data and negative data. IntFF introduces hidden units to the original Forward-Forward algorithm which are ignored in the procedure of calculating the goodness functions, or local losses, and in the procedure of predicting the labels of images. These hidden units can accomplish various mechanisms like convolutions or attention thus making the forward-forward algorithm capable of harnessing more various network structures.\\\\
Compared with the original Forward-Forward, the Integrated Forward-Forward algorithm uses shallow backpropagation to calculate the gradients in the hidden units. This one to three length of gradient information passing does not violate the locality of cortex learning that many researchers consider as the main reason why BP is not biologically plausible. Actually, even in the original Forward-Forward algorithm, we get the gradients of the parameters from the activations of the next layer, which can also be considered as shallow backpropagation of one layer.
\subsection{Relationship Between the IntFF and the Backpropagation with Stop\_Gradient Mechanism}
Compared with the backpropagation, IntFF does not have a global loss function and uses multiple local losses instead. These local losses decomposes the global loss function to different local parts of the neural network. Parameters are updated using gradients calculated from these local losses instead of a global loss function. It also uses backpropagation in a its shallow hidden units with about one or two layers of gradient information passing and can support a larger range of information passing if the researcher want.\\\\
Although local losses are explicitly different from a global loss function, it is also possible to view the gradients calculated from these local losses as the same with the gradients calculated from a global loss function with backpropagation using stop\_gradient mechanism. The stop\_gradient mechanism appears in current machine learning libraries like TensorFlow and PyTorch.It is set as a property of tensors which stops the gradients from sequentially later layers to backpropagate back to sequentially earlier layers via a path passing this specific tensor if activated. \\\\
More specifically, we can set a global loss function to the neural network explicitly as the total sum of all the local losses and use backpropagation to calculate gradients for all the network parameters with the stop\_gradient property activated at all of the local loss tensors. The gradients calculated in this way should be exactly the same with the gradients calculated using the Integrated Forward-Forward Algorithm from the co-responding local losses by the parameters.\\\\
For the researchers trying to train IntFF and make comparisons with BP, although the IntFF and backpropagation with the stop\_gradient mechanism can produce same gradients under certain configurations, this does not mean that it can be replaced by backpropagation with stop\_grad because we are still using two forward passes with positive data and negative data separately each time, which is totally different compared with the forward-backward process in backpropagation. It is worth mentioning that if we consider these local losses as a global loss function with stop\_gradient activated and use positive data and negative data mixed together, we might get numerical overflow while training. This is because in the experiments we explored, these local losses are defined as the square mean of the activations, and this might encourage the gradients to be extremely large without a careful balance of positive data and negative data. 

\section{Future Works}
Although IntFF achieves better performance than the original FF in some image classification problems and showed better resilience than backpropagation dealing with noisy data sets. There are many related problems remaining unsolved in the context of this Integrated Forward-Forward Algorithm.\\\\
1. What network structures best fit the IntFF algorithm when doing classification or other machine learning tasks? With the introduction of local losses, we can actually include various network structures into the neural network to be trained with IntFF, like convolutional layers or attention layers inside the hidden units. From the literature, we know that convolution mechanisms, attention mechanisms and other specially designed mechanisms can improve the performance of a learning algorithm. There might be other special mechanisms that suit IntFF as well.\\\\
2. From the experimental results of the IntFF compared with BP in the noisy situation, we can see that the Integrated Forward-Forward algorithm has the potential to generate neural network of some satisfactory characteristics like better robustness. Are there any other more characteristics these neural networks trained with IntFF generally have?\\\\
3. What are the best activation functions and local loss functions to use? In this work, we limited our network to use ReLU as activation and the mean squares of the selected groups of neurons as the local loss functions. Different choice might lead to significantly improved results.\\\\
4. What other tasks can the IntFF deal with? In this work, we only explored the image classification task. In literature, the applications of IntFF or the original FF are also limited. What other tasks can be done with this training method and how can we approach these tasks with networks trained with IntFF? \\\\
5. Can we harness the separation of calculations of gradients in each hidden units to design better parallel computation algorithms to increase the speed of training neural networks? Because the gradient computation procedures using IntFF are more separated and have no reliance between different hidden units compared with backpropagation, it is promising that we can find parallel implementations that can improve the training efficiency.

\end{document}